\documentclass[journal]{IEEEtran}
\ifCLASSINFOpdf
\else
\fi
\usepackage{amsmath}
\usepackage{algorithmic}
\usepackage{algorithm}
\usepackage{subfigure}
\usepackage{graphicx}
\usepackage{amsmath}
\usepackage{float}
\usepackage{multicol}
\usepackage{multirow}
\usepackage{booktabs}
\usepackage{threeparttable}
\usepackage{booktabs}
\usepackage{color}
\usepackage{url}
\usepackage{listings}
\usepackage{newfloat}
\usepackage{algorithm}
\usepackage{algorithmic}
\usepackage{amsmath} 
\usepackage{amssymb}
\usepackage{color}
\usepackage{color}
\usepackage{graphicx}
\usepackage{pythonhighlight}
\begin{document}
%
% paper title
% Titles are generally capitalized except for words such as a, an, and, as,
% at, but, by, for, in, nor, of, on, or, the, to and up, which are usually
% not capitalized unless they are the first or last word of the title.
% Linebreaks \\ can be used within to get better formatting as desired.
% Do not put math or special symbols in the title.
\title{SceneDreamer360: Text-Driven 3D-Consistent Scene Generation with Panoramic \\Gaussian Splatting}
%
%
% author names and IEEE memberships
% note positions of commas and nonbreaking spaces ( ~ ) LaTeX will not break
% a structure at a ~ so this keeps an author's name from being broken across
% two lines.
% use \thanks{} to gain access to the first footnote area
% a separate \thanks must be used for each paragraph as LaTeX2e's \thanks
% was not built to handle multiple paragraphs
%

\author{Wenrui Li,
        Fucheng Cai,
        Yapeng Mi,
        Zhe Yang,
        Wangmeng Zuo,~\IEEEmembership{~Senior Member,~IEEE,}\\
        Xingtao Wang,
        Xiaopeng Fan,~\IEEEmembership{~Senior Member,~IEEE}

\thanks{This work was supported in part by the National Key R\&D Program of China (2021YFF0900500), the National Natural Science Foundation of China (NSFC) under grants 62441202, U22B2035, and the Fundamental Research Funds for the Central Universities under grants HIT.DZJJ.2024025. (Corresponding author: Xingtao Wang.)}
\thanks{Wenrui Li, Fucheng Cai, Yapeng Mi, Wangmeng Zuo, Xingtao Wang and Xiaopeng Fan are with the Department of Computer Science and Technology, Harbin Institute of Technology, Harbin 150001, China. (e-mail: liwr@stu.hit.edu.cn;2021111124@stu.hit.edu.cn;2021113246@stu.hit.edu.cn;\\wmzuo@hit.edu.cn;xtwang@hit.edu.cn;fxp@hit.edu.cn).}
\thanks{Zhe Yang is with the School of Mathematical Sciences, University of Electronic Science and Technology of China, Chengdu, Sichuan 611731, China. (e-mail: 202221110110@std.uestc.edu.cn).}
}
\markboth{Journal of \LaTeX\ Class Files,~Vol.~14, No.~8, August~2015}%
{Shell \MakeLowercase{\textit{et al.}}: Bare Demo of IEEEtran.cls for IEEE Journals}
% The only time the second header will appear is for the odd numbered pages
% after the title page when using the twoside option.
% 
% *** Note that you probably will NOT want to include the author's ***
% *** name in the headers of peer review papers.                   ***
% You can use \ifCLASSOPTIONpeerreview for conditional compilation here if
% you desire.

% If you want to put a publisher's ID mark on the page you can do it like
% this:
%\IEEEpubid{0000--0000/00\$00.00~\copyright~2015 IEEE}
% Remember, if you use this you must call \IEEEpubidadjcol in the second
% column for its text to clear the IEEEpubid mark.

% use for special paper notices
%\IEEEspecialpapernotice{(Invited Paper)}

% make the title area
\maketitle

% As a general rule, do not put math, special symbols or citations
% in the abstract or keywords.
\begin{abstract}
Text-driven 3D scene generation has seen significant advancements recently. However, most existing methods generate single-view images using generative models and then stitch them together in 3D space. This independent generation for each view often results in spatial inconsistency and implausibility in the 3D scenes. To address this challenge, we proposed a novel text-driven 3D-consistent scene generation model: SceneDreamer360. Our proposed method leverages a text-driven panoramic image generation model as a prior for 3D scene generation and employs 3D Gaussian Splatting (3DGS) to ensure consistency across multi-view panoramic images. Specifically, SceneDreamer360 enhances the fine-tuned Panfusion generator with a three-stage panoramic enhancement, enabling the generation of high-resolution, detail-rich panoramic images. During the 3D scene construction, a novel point cloud fusion initialization method is used, producing higher quality and spatially consistent point clouds. Our extensive experiments demonstrate that compared to other methods, SceneDreamer360 with its panoramic image generation and 3DGS can produce higher quality, spatially consistent, and visually appealing 3D scenes from any text prompt. Our codes are available at \url{https://github.com/liwrui/SceneDreamer360}.
\end{abstract}

% Uncomment the following to link to your code, datasets, an extended version or similar.
%
% \begin{links}
%     \link{Code}{https://aaai.org/example/code}
%     \link{Datasets}{https://aaai.org/example/datasets}
%     \link{Extended version}{https://aaai.org/example/extended-version}
% \end{links}

\section{Introduction}
The scarcity of annotated 3D text-point cloud datasets poses substantial challenges to training 3D point cloud models directly from user queries, particularly for complex 3D scenes. This difficulty primarily stems from the high costs associated with acquiring and annotating 3D data, in addition to the precision required for modeling detailed 3D environments. To address these limitations, existing approaches frequently extend advanced 2D generative models by leveraging 2D object priors as a bridge to 3D neural radiance fields. For instance, CLIP-NeRF \cite{wang2022clipnerf} uses the CLIP model \cite{radford2021learningclip} to control the shape and appearance of 3D objects based on textual prompts or images. Similarly, Dream Fields \cite{jain2022zerodreamfields} integrates neural rendering with image and text representations to produce diverse 3D objects from natural language descriptions. Dreamfusion \cite{poole2022dreamfusion} further advances this approach by optimizing NeRF without the need for 3D training data, instead relying solely on a pre-trained 2D text-to-image diffusion model for text-to-3D synthesis.

\begin{figure}
	\centering
	\includegraphics[width=1\linewidth]{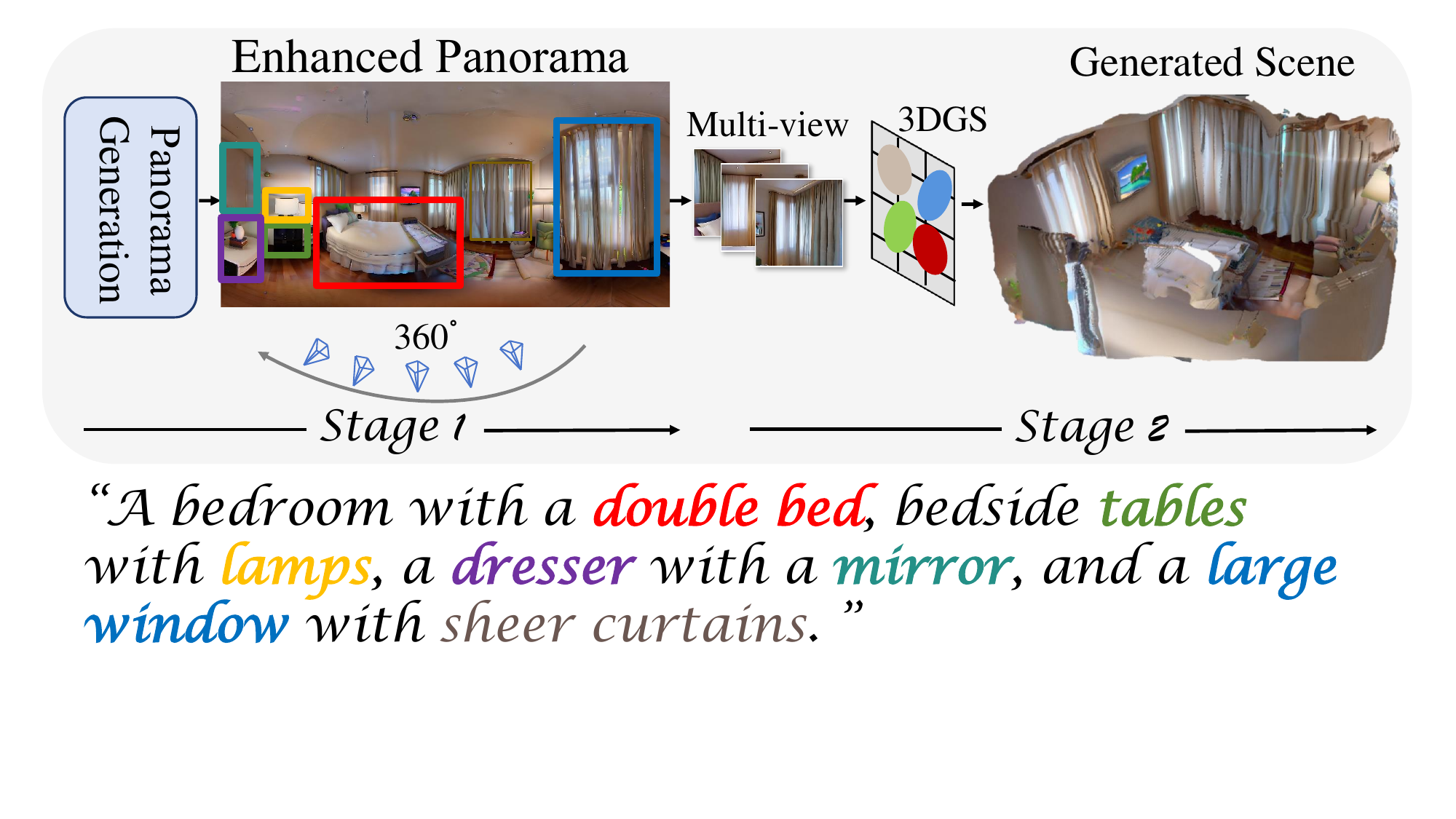}
	\caption{We introduce SceneDreamer360, a text-based 3D scene generation framework designed to create realistic 3D scenes with high consistency across different viewpoints. SceneDreamer360 consists of two stages. In the first stage, the Panorama Generation module creates enhanced panoramic images to provide a complete and consistent spatial prior for the 3D scene. In the second stage, 3D Gaussian Splatting is used to reconstruct multi-view spatial images, resulting in a complete and spatially consistent point cloud.
}
	\label{fig1}
\end{figure}

Despite significant advancements in NeRF-based point cloud generation methods, achieving consistent, fine-grained point cloud details remains a challenge. For example, the Text2Room method \cite{hollein2023text2room} often produces point clouds that are discontinuous, lack detailed features, and require prolonged rendering times. Similarly, LucidDreamer \cite{chung2023luciddreamer} generates panoramic point clouds with numerous inconsistencies, resulting in incomplete and fragmented panoramas. PanFusion \cite{panfusion2024} also struggles with generating high-quality panoramic images from complex, extended text inputs, frequently yielding blurred areas and deformed objects. Moreover, PanFusion’s panoramic images are relatively low-resolution ($512 \times 1024$), leading to a reduced spatial-to-pixel point ratio, which further contributes to blurred, suboptimal point cloud rendering. To address these challenges, we propose incorporating 3D Gaussian Splatting (3DGS) \cite{kerbl20233dgs} for multi-scene generation, specifically aimed at producing more finely detailed and consistent complex scene point clouds. By employing a Gaussian distribution for scene modeling, 3DGS enables more precise representation of complex structures and improves reconstruction accuracy. Additionally, 3DGS’s optimized rendering algorithm significantly reduces rendering times while preserving high quality, resulting in faster and more efficient generation of detailed point clouds.

In this paper, we introduce SceneDreamer360, a novel framework for text-driven 3D-consistent scene generation using panoramic Gaussian splatting (3DGS). Our method operates in two stages, as illustrated in Fig. \ref{fig1}: first, we enhance panoramic images to establish 3D consistency priors, and second, we apply 3DGS to reconstruct high-quality, text-aligned point clouds. This dual-stage approach ensures both accurate 3D structure and enhanced visual quality and realism in the generated scenes. During the panorama generation and enhancement stage, existing techniques frequently rely on iterative, progressive scene rendering, which can lead to consistency issues across successive renderings. To address this challenge, we enhance the PanFusion model \cite{panfusion2024} by integrating a multi-layer perceptron (MLP) \cite{murtagh1991multilayer} and a LoRA layer \cite{hu2021lora} into its final processing stage. These additions, combined with training on the Habitat Matterport Dataset \cite{ramakrishnan2021habitat}, produce a model checkpoint fine-tuned to the specific demands of panoramic image generation. This enhancement step produces high-quality panoramas, forming a solid foundation for subsequent point cloud rendering. Further, we upscale and refine the generated panoramas using ControlNet (CN) \cite{zhang2023addingcontrolnettile} and RealESRGAN \cite{wang2021realesrgan} techniques, achieving resolutions up to 6K. This upscaling process preserves fine details and enhances image fidelity, which is crucial for realistic point cloud rendering. High-resolution panoramas contribute to more detailed and visually appealing scenes, which are important for achieving photorealism in the final output. To optimize the point cloud rendering process, we introduce a novel point cloud initialization method that improves 3D consistency and reduces rendering time. This step ensures better alignment of the resulting point clouds with the panoramic views, creating a more cohesive and immersive 3D experience. Additionally, the robust 3D representation capabilities of 3DGS enable the creation of complete, high-quality point cloud images that are consistent with the input text prompts. Our contributions can be summarised as follows:
\begin{itemize}
    \item We propose SceneDreamer360, a domain-free high-quality 3D scene generation method, achieving better domain generalization in 3D scene generation by leveraging the panorama generation and 3D Guassion Splatting.
    \item Our method adapts panorama generation to provide a consistency prior for scenes, fine-tunes the model, and applies a three-stage enhancement to add more detail and spatial consistency to the final scene generation.
    \item To improve scene reconstruction, we employed 3D Gaussian Splatting and introduced a novel point cloud initialization and fusion algorithm, resulting in enhanced performance of the final generated point cloud scenes.
\end{itemize}

The remainder of this paper is organized as follows: Section \ref{relatedwork} reviews the background and related work on 3D scene generation, 3D scene representation, and panorama generation. Section \ref{proposedmethod} describes the architecture and functionality of the proposed Scenedreamer360 model in detail. Section \ref{experiments} presents the experimental results and visualizations, highlighting the effectiveness of our approach. Finally, Section \ref{conclusion} summarizes the study's main findings and contributions.

\section{Related Work}\label{relatedwork}
\subsection{3D Scene Generation} The field of 3D scene generation has evolved significantly, drawing inspiration from various breakthroughs in image generation techniques. Early approaches leveraged Generative Adversarial Networks (GANs) \cite{goodfellow2014generativeadversarialnetworks}, attempting to create multi-view consistent images \cite{chan2021piganperiodicimplicitgenerative,niemeyer2021girafferepresentingscenescompositional} or directly generate 3D representations like voxels \cite{nguyen2019hologan,nguyen2020blockgan} and point clouds \cite{achlioptas2018learning,shu20193d}. However, these methods were hindered by GAN's inherent learning instability and the memory constraints of 3D representations, limiting the quality of generated scenes.

The advent of diffusion models \cite{ho2020denoising19,song2020denoising47,tcsvt3} and their success in image generation \cite{ramesh2021zero37,rombach2022high39} sparked a new wave of research in 3D scene generation. Researchers began applying diffusion models to various 3D representations, including voxels \cite{zhou20213d60}, point clouds \cite{vahdat2022lion59}, and implicit neural networks \cite{poole2022dreamfusion34}. While these approaches showed promise, they often focused on simple, object-centric examples due to their inherent nature.

To address more complex scenarios, some generative diffusion models employed meshes as proxies, diffusing in UV space. This approach enabled the creation of large portrait scenes through continuous mesh building \cite{fridman2024scenescape13} and the generation of indoor scenes \cite{cohen2023set8,lei2023rgbd22} and more realistic objects \cite{qian2023magic1235}. Currently, an increasing number of studies \cite{chung2023luciddreamer,zhou2024dreamscene360,ma2024fastscene,li2024art3d} have incorporated 3DGS \cite{kerbl20233dgs} as a 3D representation to complement diffusion models in generating more complex 3D scenes. For instance, LucidDreamer \cite{chung2023luciddreamer} pioneered the integration of 3D Gaussian Splatting into scene generation, demonstrating the efficacy of 3DGS. RealmDreamer \cite{shriram2024realmdreamer} optimizes a 3D Gaussian Splatting representation to align with complex text prompts. However, due to diffusion models generating one viewpoint at a time before merging and elevating to 3D space, the resulting 3D scenes often lack realism and consistency. To address this issue, some recent methods such as DreamScene360 \cite{zhou2024dreamscene360} and FastScene \cite{ma2024fastscene} have introduced panoramic images to generate more coherent 3D scenes. Nevertheless, these approaches still produce scenes lacking in detail and quality, primarily due to limitations in panoramic image generation models and view fusion algorithms.

In contrast, to address the issues of quality and spatial consistency, our method leverages the promising 3D Gaussian Splatting technique and implements a three-stage enhancement process for the generated panoramic images. Concurrent work HoloDreamer \cite{zhou2024holodreamer} has explored similar concepts. However, our approach is distinguished by an innovative design in the panoramic image enhancement stage and a novel point cloud fusion algorithm. Our approach improves the overall quality and spatial coherence of the generated 3D scenes, advancing the state-of-the-art in text-driven 3D scene generation.

\subsection{3D Scene Representation} The field of 3D scene representation \cite{tcsvt4,tcsvt5} has evolved significantly, encompassing a variety of approaches with distinct trade-offs and applications. Traditional explicit methods such as point clouds, meshes, and voxels have long been fundamental to 3D modeling, offering intuitive control and fast rendering through rasterization. However, these methods often require a large number of elements to achieve high resolution, which can be computationally expensive. To address this limitation, more complex primitives are developed, including cuboids \cite{tulsiani2017learning52,tcsvt1,tcsvt2} and polynomial surfaces \cite{yavartanoo20213dias56}, offering more efficient expression of geometry but still struggling with realistic color representation.

Recent advancements have led to the emergence of implicit neural representations, such as Signed Distance Functions (SDF) \cite{takikawa2021neural49} and Neural Radiance Fields (NeRF) \cite{mildenhall2021nerf26}. These methods, particularly NeRF, have demonstrated the ability to represent complex 3D shapes and textures with rich details. However, they face challenges in terms of handling and optimization efficiency. To improve upon these limitations, subsequent research has focused on combining volume rendering with explicit representations, utilizing structures like sparse voxels \cite{fridovich2022plenoxels14,liu2020neural24,sun2022direct48,yu2021plenoctrees58}, featured point clouds \cite{xu2022point55}, Multi-Level Hierarchies \cite{muller2021real27,muller2022instant28}, tensors \cite{chen2022tensorf5}, infinitesimal networks \cite{garbin2021fastnerf15,reiser2021kilonerf38}, triplanes \cite{chan2022efficient4}, and polygons \cite{chen2023mobilenerf7}.

Among these advancements, 3D Gaussian Splatting(3DGS) \cite{kerbl20233dgs} has emerged as a particularly promising approach. It represents complete and unbounded 3D scenes by effectively 'splatting' Gaussians, utilizing spherical harmonics and opacity for strong representational capabilities. 3DGS balances quality and efficiency by supporting alpha-blending and differentiable rasterization, enabling fast 3D scene optimization. It refines point clouds details through a split-and-clone mechanism, making it ideal for complex scene generation where scene bounds are uncertain.
% It starts with point clouds and refines details through a split-and-clone mechanism, making it ideal for complex scene generation where scene bounds are uncertain.
\begin{figure*}[ht]
	\centering
	\includegraphics[width=1\linewidth]{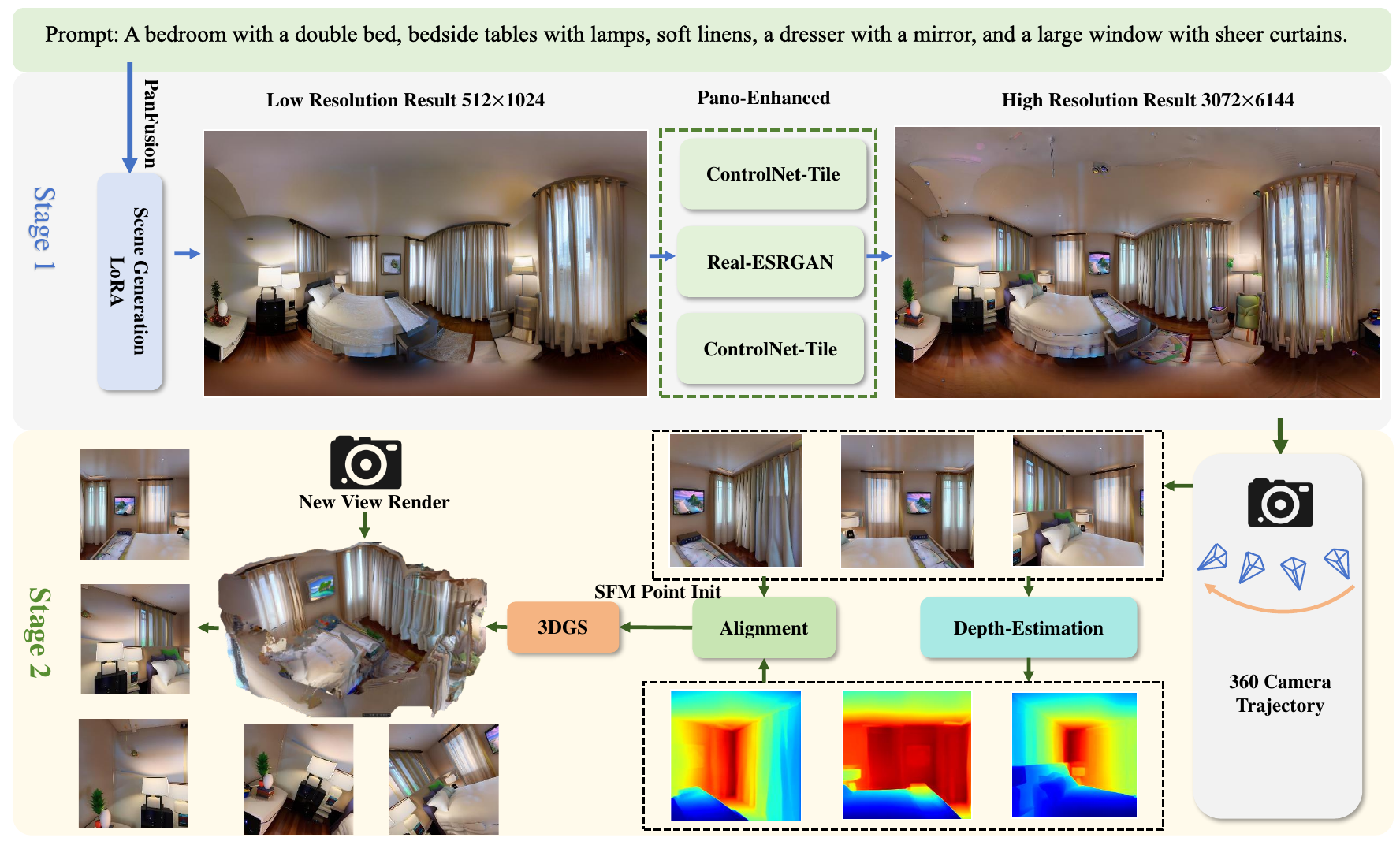}
	\caption{The architecture of the SceneDreamer360. SceneDreamer360 generates an initial panorama from any open-world textual description using the fine-tuned PanFusion model. This panorama is then enhanced to produce a high-resolution $3072 \times 6144$ image. In the second stage, multi-view algorithms generate multi-view images, and a monocular depth estimation model provides depth maps for initial point cloud fusion. Finally, 3D Gaussian Splatting is used to reconstruct and render the point cloud, resulting in a complete and consistent 3D scene.
}
	\label{fig2}
\end{figure*}

\subsection{Panorama Generation} Panoramas offer a comprehensive, unobstructed view that captures extensive scene areas. Recent advancements in panorama generation have significantly improved the quality and consistency of 360-degree images. While early attempts like PanoGen \cite{li2024panogen} and MultiDiffusion \cite{bar2023multidiffusion} used pre-trained diffusion models to create long images from text prompts, these methods often resulted in inconsistencies and lacked true panoramic properties. To address these limitations, researchers have developed more sophisticated approaches. MVDiffusion \cite{Tang2023mvdiffusion} introduced a correspondence-aware attention mechanism to generate consistent multi-view images simultaneously. This method ensures that the different views maintain coherence and alignment, which is crucial for creating seamless panoramic images. However, MVDiffusion still faced challenges in achieving perfect continuity at the image edges. StitchDiffusion \cite{wang2024customizing38} improved upon it by using LoRA \cite{hu2021lora} fine-tuning to generate complete 360-degree panoramas with enhanced continuity between image edges. This approach leverages the Low-Rank Adaptation (LoRA) technique to fine-tune the model, ensuring that the transitions between different parts of the panorama are smooth and natural. PanFusion \cite{panfusion2024} proposed a novel dual-branch diffusion model to generate a 360-degree image from a text prompt. This model utilizes two separate branches to handle different aspects of the image generation process, allowing for more detailed and context-aware outputs. The dual-branch architecture helps in capturing both the global structure and local details of the scene, resulting in high-quality panoramic images.

In this paper, we incorporated MLP and LoRA into the PanFusion model and performed fine-tuning to enhance the consistency of the generated panoramic images. By integrating a Multi-Layer Perceptron (MLP) with LoRA in the PanFusion framework, we aim to further improve the coherence and quality of the panoramas. The MLP component helps in refining the details and ensuring that the generated images are not only consistent but also rich in detail. Furthermore, recent studies have explored the integration of machine learning techniques with traditional computer vision methods to further enhance panorama generation. For instance, hybrid models that combine deep learning with geometric alignment techniques have shown promise in reducing artifacts and improving the overall visual quality of the panoramas. These advancements highlight the ongoing efforts to push the boundaries of what is possible in panorama generation, making it an exciting and rapidly evolving field. Overall, the integration of advanced attention mechanisms, fine-tuning techniques, and hybrid models has significantly advanced the state of panorama generation, enabling the creation of high-quality, seamless 360-degree images from text prompts.

\section{Method}\label{proposedmethod}
This method introduces a novel pipeline for generating high-resolution, 360-degree panoramic point cloud scenes from arbitrary spatial text inputs. The pipeline consists of two main stages: Panoramic Image Generation and Processing and Point Cloud Construction with Gaussian Splatting. As illustrated in Fig.~\ref{fig2}, the SceneDreamer360 framework effectively combines these stages to produce visually coherent and spatially consistent 3D scenes, capturing fine details and maintaining high image fidelity.

In the first stage, an adapted PanFusion \cite{panfusion2024} model generates initial panoramic images, which are subsequently refined through a three-step super-resolution process, enhancing both image details and resolution. Next, we use the high-resolution panorama as input and apply the Equirectangular to Perspective (e2p) algorithm to obtain multiple scene images from different perspectives. In the second stage, point clouds are constructed using a 3D Gaussian Splatting \cite{kerbl20233dgs} technique. This process involves initializing the point cloud, applying masking to remove duplicates for a more complete map, and training a 3D Gaussian splatting model along a new rendering trajectory to produce a detailed 3D spatial scene point cloud.

\subsection{Panoramic Image Generation and Processing}

The goal of Panoramic Image Generation and Optimization is to convert a long-text description of a complex scene into high-quality 360-degree panoramas with high resolution. The specific steps are as follows:

\subsubsection{Complex Long-Text Generation}
We used the GPT-4 model to generate a series of similar complex scene descriptions to increase the diversity and richness of the training data. The user first inputs an initial long text description of a complex scene. To generate more texts with similar semantics and content, we utilize GPT-4 to expand and transform it. These generated texts cover different details and perspectives, making the subsequent panorama generation more diverse and comprehensive.

\subsubsection{Finetune PanFusion Model}
Given the input long text representation $T$ and its embedding $e_T$, the representation after the MLP layer is:
\begin{equation}
\mathbf{h} = \sigma(\mathbf{W}_2 \cdot \text{ReLU}(\mathbf{W}_1 \cdot \mathbf{e}_T + \mathbf{b}_1) + \mathbf{b}_2),
\end{equation}
where $W_1$ and $W_2$ are the weight matrices of the MLP layer, $b_1$ and $b_2$ are the bias terms, and $\sigma$ is the activation function.

\subsection{Panoramic Image Generation}
Our method utilized the PanFusion model \cite{panfusion2024} to generate panoramas. However, because the PanFusion model was originally trained on simple short text prompts, it required appropriate adjustments and fine-tuning to handle complex long texts. Therefore, we added a Multi-Layer Perceptron (MLP) \cite{murtagh1991multilayer} layer and a Low-Rank Adaptation (LoRA) \cite{hu2021lora} layer to the original PanFusion model, fine-tuning it using the reannotated Habitat Matterport Dataset \cite{ramakrishnan2021habitat}. The MLP layer captures higher-order features in complex long texts, while the LoRA layer enhances the model's representation ability through low-rank matrix decomposition. For model fine-tuning, we first reannotated the dataset using the generated complex long texts, pairing them with corresponding panoramas to form new training samples. Next, we froze the pre-trained parts of the PanFusion model and trained only the newly added MLP and LoRA layers. This ensures that the model effectively utilizes information from complex long texts to generate high-quality panoramas. These panoramas encapsulate the details and scenes described in the input text, serving as foundational data for subsequent steps.

Given the input long text representation $T$ and its embedding $e_T$, the representation after the MLP layer is:
\begin{equation}
\mathbf{h} = \sigma(\mathbf{W}_2 \cdot \text{ReLU}(\mathbf{W}_1 \cdot \mathbf{e}_T + \mathbf{b}_1) + \mathbf{b}_2),
\end{equation}
where $W_1$ and $W_2$ are the weight matrices of the MLP layer, $b_1$ and $b_2$ are the bias terms, and $\sigma$ is the activation function.

The representation of the LoRA layer is:
\begin{equation}
    \mathbf{h}' = \mathbf{h} + \mathbf{U} \cdot \text{softmax}(\mathbf{V} \cdot \mathbf{h}),
\end{equation}
where $U$ and $V$ are the low-rank matrices of the LoRA layer.

\subsubsection{Dataset Reannotation} We reannotated the dataset using the generated complex long texts. Specifically, we paired the complex long texts with corresponding panoramas to form new training samples.

\subsubsection{Model Fine-tuning} During fine-tuning, we froze the pre-trained parts of the PanFusion model and only trained the newly added MLP and LoRA layers. This ensures that the model can effectively utilize information from the complex long texts to generate high-quality panoramas.

The adjusted and fine-tuned PanFusion model is applied to the generated long text collection, producing 360-degree panoramas. These panoramas contain details and scenes described in the input text and serve as the foundational data for subsequent steps.

\subsection{Panoramic Image Optimization} As the pre-trained PanFusion model can only generate panoramas with a resolution of $512\times1024$, directly using these panoramas for multi-view extraction and point cloud reconstruction would result in sparse point clouds with blurry 3D Gaussian rendering \cite{kerbl20233dgs}. To meet the requirements of point cloud reconstruction, it is necessary to further enhance the resolution and quality of the panoramas. The specific steps are as follows: We adopted a three-step super-resolution reconstruction method involving ControlNet-Tile \cite{zhang2023addingcontrolnettile}, RealESRGAN \cite{wang2021realesrgan}, and a second application of ControlNet-Tile to maximize the enhancement of image details and resolution.

\subsubsection{ControlNet-Tile Super-Resolution Reconstruction} 
The ControlNet-Tile model divides the panorama into distinct tiles through semantic recognition. Each tile is analyzed by the model to determine which details to add and which parts to repair. Based on tiling, ControlNet-Tile enhances the details of each tile. The model can identify various areas in the image and enhance the details of each area, adding additional detail where necessary. The detail enhancement process can be expressed as:
\begin{equation}
    \mathbf{I}_{enhanced} = f_{tile}(\mathbf{I}_{ori},t),
\end{equation}
where $f_{tile}$ represents the ControlNet-Tile model, $I_{ori}$ represents the original panorama, and $t$ represents the input text.

\subsubsection{RealESRGAN Super-Resolution Reconstruction} 
Based on ControlNet-Tile, we further employed RealESRGAN for higher precision super-resolution reconstruction. RealESRGAN is an advanced super-resolution method based on generative adversarial networks, capable of enhancing resolution while preserving natural details and visual consistency. The super-resolution reconstruction process can be expressed as:
\begin{equation}
    \mathbf{I}_{super} = g_{ESRGAN}(\mathbf{I}_{enhanced}),
\end{equation}
where $g_{ESRGAN}$ represents the RealESRGAN model.

\subsubsection{ControlNet-Tile Re-Optimization} After RealESRGAN processing, we applied ControlNet-Tile once more for additional refinement and optimization. The purpose of this second application of ControlNet-Tile is to further enhance the details and quality of the image at the increased resolution. The final optimized image $I_final$ is represented as:
\begin{equation}
    \mathbf{I}_{final} = f_{tile}(\mathbf{I}_{super},t).
\end{equation}

By following these steps, we increased the panorama's resolution to $3072\times6144$. This high-resolution panorama preserves the original image's overall structure while significantly enhancing details and clarity, thus providing high-quality input data for subsequent point cloud reconstruction.

\subsection{Multi-View Image Acquisition} 
After obtaining the high-resolution panorama, we need to extract images from specific viewpoints for point cloud reconstruction. We select a suitable set of camera poses $P_i \in \mathbb{R}^{4 \times 4}$ based on the requirements of point cloud reconstruction. The selection of these camera poses is based on geometric and visual coverage principles, ensuring that the chosen perspectives comprehensively cover important details and areas of the panorama. Specifically, we use an optimization-based viewpoint selection algorithm that considers the distribution of viewpoints, parallax effects, and coverage of reconstruction areas. The camera pose $P_i$ is represented as:
\begin{equation}
    \mathbf{P}_i = \begin{bmatrix}
\mathbf{R}_i & \mathbf{T}_i \\
\mathbf{0} & 1
\end{bmatrix}.
\end{equation}

The rotation matrix $R_i$ is expressed as:
\begin{equation}
    \mathbf{R}_i = \begin{bmatrix}
\cos(\theta) & 0 & \sin(\theta) \\
0 & 1 & 0 \\
-\sin(\theta) & 0 & \cos(\theta)
\end{bmatrix}.
\end{equation}

The translation matrix $T_i$ is expressed as:
\begin{equation}
    \mathbf{T}_i = \begin{bmatrix}
T_x \\
T_y \\
T_z
\end{bmatrix}
=\begin{bmatrix}
0 \\
0 \\
0
\end{bmatrix}.
\end{equation}

% In our method, the translation matrix is set to zero, enabling the camera to rotate in a fixed position.

By using the Equirectangular to Perspective (e2p) algorithm, we transformed the panorama to generate perspective images from specific viewpoints. The e2p algorithm converts spherical panoramas into multiple planar images from different viewpoints, thereby creating parallax between views. The perspective transformation formula is:
\begin{equation}
    \mathbf{I}_{perspective}(\mathbf{u}, \mathbf{v}) = \mathbf{E}(\mathbf{P}(f(\mathbf{u}, \mathbf{v}),\mathbf{R}_i,\mathbf{T}_i),\mathbf{S}),
\end{equation}
where $f(\mathbf{u}, \mathbf{v})$ represents the mapping function from planar coordinates $(\mathbf{u}, \mathbf{v})$ to spherical coordinates, $\mathbf{S}$ represents the specified output size, $\mathbf{E}$ and represents the view size transformation function.

Based on the camera poses, the e2p algorithm generates perspective images from the panorama. These perspective images retain the high resolution and details of the panorama while introducing parallax between views, providing rich perspective information for subsequent point cloud reconstruction.

By employing these methods, we can effectively generate high-quality, 360-degree panoramas directly from text prompts. This approach provides high-resolution and detailed input data, forming an optimal basis for the subsequent point cloud reconstruction. Using precise perspective transformations, we produce a series of viewpoint images that are well-suited to the requirements of point cloud reconstruction. These images not only capture various angles but also retain the visual fidelity and spatial coherence needed for constructing accurate 3D scenes. Consequently, they serve as a solid foundation for the point cloud reconstruction process, enhancing the realism and consistency of the final generated scene.

\subsection{Point Cloud Construction with Gaussian Splatting}

After obtaining high-quality multi-view panoramic images, we apply the 3D Gaussian Splatting method for point cloud generation. We directly use 3D Gaussians to represent point clouds and have designed a novel point cloud alignment method. Additionally, we improved the point cloud initialization by applying masking to remove duplicates, resulting in a more refined and complete initial point cloud map. The point cloud reconstruction process is divided into two phases: Point Cloud Initialization and Rendering.

\subsubsection{Point Cloud Initialization} The initial point cloud must be generated from multi-view images to enable reconstruction using 3D Gaussian Splatting \cite{kerbl20233dgs}. To generate a 360-degree spatial point cloud from multi-view images, a rendering trajectory centered around the observation viewpoint is required. This trajectory should cover the bottom, middle, and top of the spatial area. Using the LucidDreamer \cite{chung2023luciddreamer} framework, we incorporate additional camera viewpoints to create a more accurate spatial point cloud. For the set of multi-view images $\mathbf{I_1},\mathbf{I_2},...,\mathbf{I_n}$ derived from the panoramic image, each image $\mathbf{I_i}$ has a corresponding position $\mathbf{P_i}$. Using the monocular depth estimation model ZoeDepth \cite{bhat2023zoedepth}, we measure the depth map $\mathbf{D_i}$ for each $\mathbf{I_i}$, forming the depth map set {$\mathbf{D_1},\mathbf{D_2},...,\mathbf{D_n}$}. We initialize the point cloud using a cyclic generation approach. The camera's intrinsic matrix and the extrinsic matrix of $\mathbf{I_0}$ are denoted as $\mathbf{K}$ and $\mathbf{P_0}$, respectively. The initial point cloud for a single viewpoint is derived using the following formula:
\begin{equation}
\rho_0 = \tau_{2 \rightarrow 3} (\mathbf{I}_0, \mathbf{D}_0, \mathbf{K}, \mathbf{P}_0),
\end{equation}
where $\tau_{2 \rightarrow 3} (\cdot)$ is a function that elevates a 2D image to a 3D point cloud. This function generates an initial point cloud $\rho_i$, corresponding to a viewpoint $P_i$.

For point cloud fusion from different perspectives, we employed a masking method to remove duplicates. Specifically, we defined a mask \( \mathbf{M_i} \) for each generated \(\rho_i\), where the mask projects the already generated \(\rho_{i-1}\) onto the 2D image space through the corresponding view \( \mathbf{P_i} \). The projected part of the mask has values greater than 0, while the rest remains black with a value of 0. At the final stage of generation, we remove the duplicated parts that have already been projected, effectively merging the point clouds from different perspectives and reducing duplicate points. Meanwhile, we estimate the optimal depth scaling factor \( d_i \) to minimize the distance between the 3D points of the new image and the corresponding points in the original point cloud \( \mathbf{P_{i-1}} \). Then, by multiplying the factor \( d_i \) with the measured depth map \( \mathbf{D_i} \), we calculate the corrected depth map \( d_i\mathbf{D_i} \), which is subsequently used for the generation of \(\rho_i\). The complete process is illustrated in the following formula:
\begin{equation}
d_i = \mathop{\text{argmin}}_{d} \left( \sum \left\| \tau_{2 \rightarrow 3}\left( [\mathbf{I}_i, d \mathbf{D}_i], \mathbf{K}, \mathbf{P}_i \right) - \rho_{i-1} \right\|_1 \right),
\end{equation}
\begin{equation}
\rho_i = \tau_{2\rightarrow 3} (\mathbf{I}_i-\mathbf{M}_i,d_i\mathbf{D}_i,\mathbf{K},\mathbf{P}_i),i\ne 0.
\end{equation}

Finally, through continuous iteration of the formula, we fuse the point clouds to form the final point cloud \(\Omega\). 
\begin{equation}
\Omega  =  {\textstyle \bigcup_{i=1}^{|P|}\varphi \left \{\rho_i,\rho_{i-1} \right \}}.
\end{equation}

The function \(\varphi\) represents the depth alignment function, and all \(\rho\) are integrated to obtain the complete initial point cloud \(\Omega\). \textbf{The overall algorithm is shown in Algorithm 1.}
\begin{algorithm}[tb]
\caption{Point cloud initialization algorithm}
\label{alg:algorithm}
\textbf{Input}: Panoramic multiview RGBD images [$\mathbf{I}_i$,$\mathbf{D}_i$]$_{i=0}^{N}$\\
\textbf{Input}: Camera intrinsic $\mathbf{K}$, extrinsics $\left \{ \mathbf{P}_i \right \} _{i=0}^{N} $\\
\textbf{Output}: Complete point cloud $\Omega$

\begin{algorithmic}[1] %[1] enables line numbers
\STATE $\rho_0 = \tau_{2\rightarrow 3} (\mathbf{I}_0,\mathbf{D}_0,\mathbf{K},\mathbf{P}_0)$
\FOR{$i = 1$ \TO $N$}
\STATE $d_i = 1$
\STATE $\mathbf{M}_i = \psi_{3\to 2} (\rho_{i-1},\mathbf{K},\mathbf{P}_i)$
\WHILE{\textit{not converged}}
\STATE $\tilde{\rho_i}  = \tau_{2\rightarrow 3} (\mathbf{I}_i,d_i\mathbf{D}_i,\mathbf{K},\mathbf{P}_i)$
\STATE $\mathcal{L}_{d} \leftarrow \frac{1}{\left\|\mathbf{M}_{i}=1\right\|} \sum_{\mathbf{M}_{i}=1}\left\|\tilde{\rho_i}-\rho_{i-1}\right\|_{1}$
\STATE Calculate $\nabla_{d}\mathcal{L}_{d}$
\STATE $d_{i} \leftarrow d_{i}-\alpha \nabla_{d} \mathcal{L}_{d}$
\ENDWHILE
\STATE $\rho_i = \tau_{2\rightarrow 3} (\mathbf{I}_i-\mathbf{M}_i,d_i\mathbf{D}_i,\mathbf{K},\mathbf{P}_i)$
\STATE $\Omega  =  \rho_{i-1}\bigcup \rho_{i}$
\ENDFOR
\STATE \textbf{return} $\Omega$
\end{algorithmic}
\end{algorithm}

\subsubsection{Rendering with Gaussian Splatting} 3D Gaussian Splatting is a technique that efficiently models 3D scenes by leveraging a fast and accurate point-based representation. Following the generation of the initial point cloud, we use this point cloud to train a 3D Gaussian Splatting model. The generated point cloud, denoted as \(\Omega\) , serves as the initial Structure from Motion (SfM) points, which accelerates network convergence and encourages the model to focus on producing detailed and realistic representations of the scene. To train the model, we utilize multi-view image slices as training images and incorporate the image view augmentation technique from LucidDreamer to enrich the data. Since the input multi-view images may not fully cover the entire 3D space, we project the \(\Omega\) point cloud into 2D from various camera viewpoints, effectively increasing the number of supervisory images and expanding spatial coverage. This augmentation process helps to provide more comprehensive training data, ensuring better spatial consistency and detail across the 3D scene.

\begin{equation}
\mathbf{I}_i, \mathbf{M}_i = \psi_{3 \to 2} (\Omega, \mathbf{K}, \mathbf{P}_i),
\end{equation}
where \(\psi_{3 \to 2} (\cdot)\) is the function that projects the 3D point cloud onto the 2D plane. Through this method, we achieve a more consistent and complete scene point cloud.

\begin{figure*}
	\centering
	\includegraphics[width=1\linewidth]{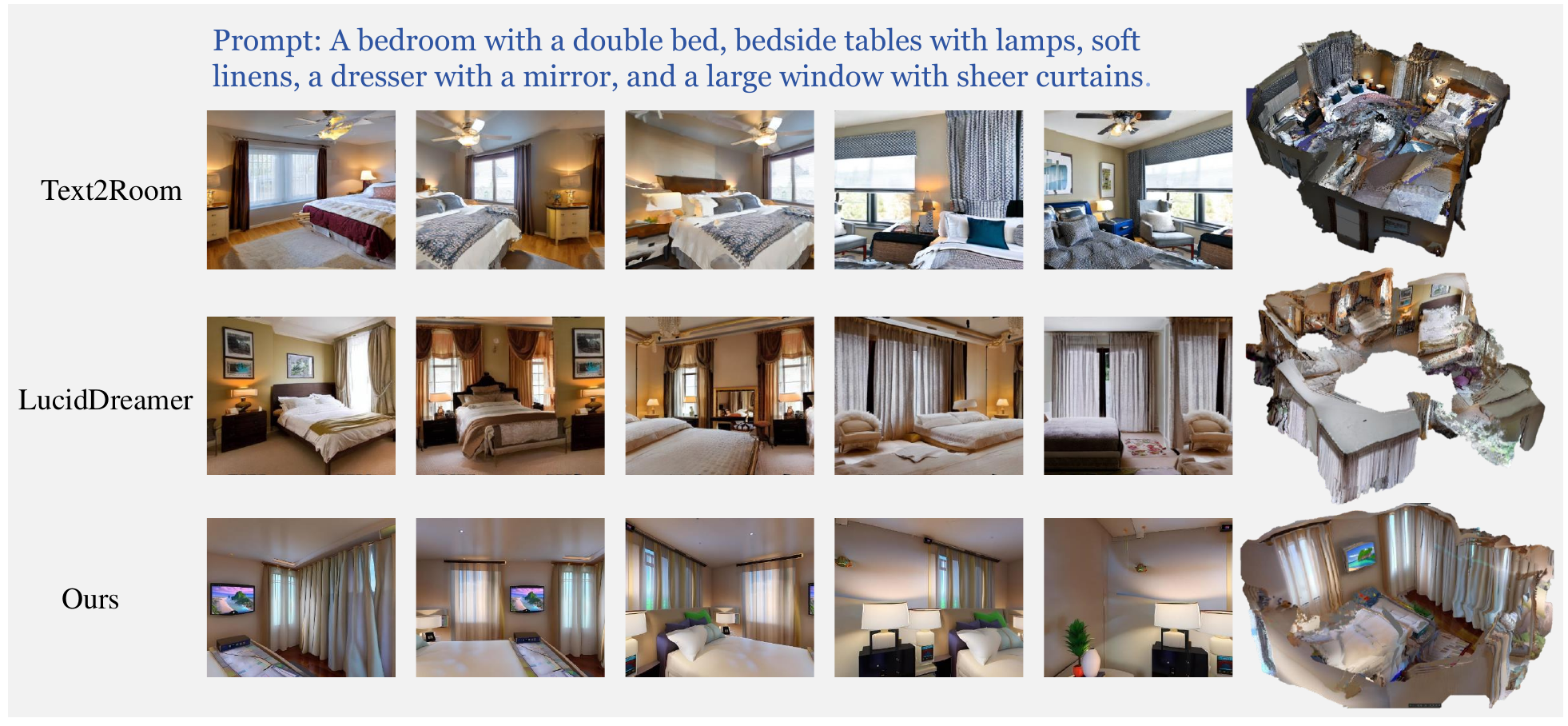}
	\caption{The visualization comparison of the proposed SceneDreamer360 between current methods. The images on the left display the new viewpoint renderings, demonstrating how each method handles various perspectives. In contrast, the images on the right present the spatial views of the generated 3D scenes, allowing a direct comparison of the spatial consistency and overall structure of the reconstructions.}
	\label{compare_fig4}
\end{figure*}
\section{Experiments}\label{experiments}
In this section, we conducted a thorough and comprehensive evaluation of our method to validate its superiority. We performed both qualitative and quantitative evaluations that our method can generate clearer and higher quality 3D scenes.

\subsection{Implementation Details} In the first stage, we generate high-quality panoramic images using a modified PanFusion model, which has been fine-tuned on the Habitat Matterport Dataset. To enhance its capacity for producing realistic 3D panoramic images, we incorporate MLP and LoRA layers into the model. This fine-tuning enables the generation of more visually coherent and lifelike panoramas. We further apply a three-step super-resolution process, using ControlNet-Tile and RealESRGAN, which enhances the resolution of these panoramas to $3072\times6144$, preserving fine details and enriching the image quality for subsequent stages.

In the second stage, we adapt and modify the point cloud reconstruction component from LucidDreamer. This adaptation allows us to effectively convert the high-resolution panoramic images into detailed 3D point clouds, ensuring consistency with the input scene and text prompt. This two-stage process of panorama generation and point cloud reconstruction is designed to maximize visual fidelity and structural accuracy in the final 3D scenes.

Our method is implemented using PyTorch. We obtain camera viewpoints from given camera poses and use the e2p algorithm to extract corresponding multi-view images from the panorama. The specific implementation method is as follows:

\definecolor{dkgreen}{rgb}{0,0.6,0}
\definecolor{gray}{rgb}{0.5,0.5,0.5}
\definecolor{mauve}{rgb}{0.58,0,0.82}
\lstset{frame=tb,
  language=Python,
  aboveskip=3mm,
  belowskip=3mm,
  showstringspaces=false,
  columns=flexible,
  basicstyle={\tiny\rmfamily},
  numbers=left,%设置行号位置none不显示行号
  %numberstyle=\tiny\courier, %设置行号大小
  numberstyle=\tiny\color{gray},
  keywordstyle=\color{blue},
  commentstyle=\color{dkgreen},
  stringstyle=\color{mauve},
  breaklines=true,
  breakatwhitespace=true,
  escapeinside=``,
  tabsize=4,
  extendedchars=false
}

\inputpython{test.py}{1}{12}

For evaluation, we generate a diverse set of prompts using GPT-4, covering a broad spectrum of scene types to comprehensively test our model's versatility. We then assess the quality of our outputs using multiple metrics, including CLIP-Score, PSNR, and others, to evaluate both visual similarity and fidelity to the input text prompts. All experiments are conducted on an NVIDIA A100-80G GPU, ensuring sufficient computational power for handling high-resolution image processing and point cloud generation.

% We compared SceneDreamer360 with existing methods, Text2Room and LucidDreamer.
\begin{figure*}
	\centering
	\includegraphics[width=1\linewidth]{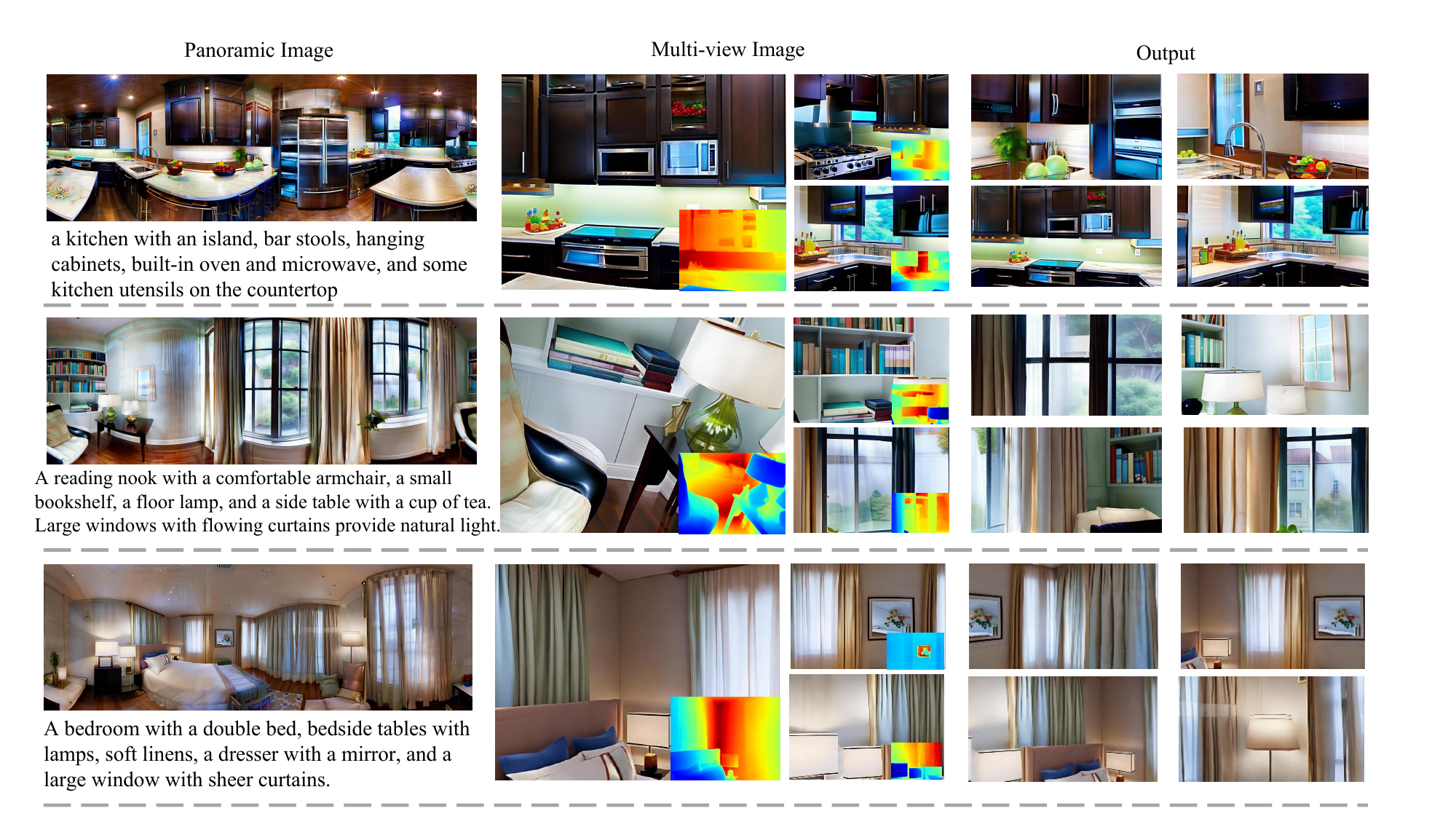}
        \includegraphics[width=1\linewidth]{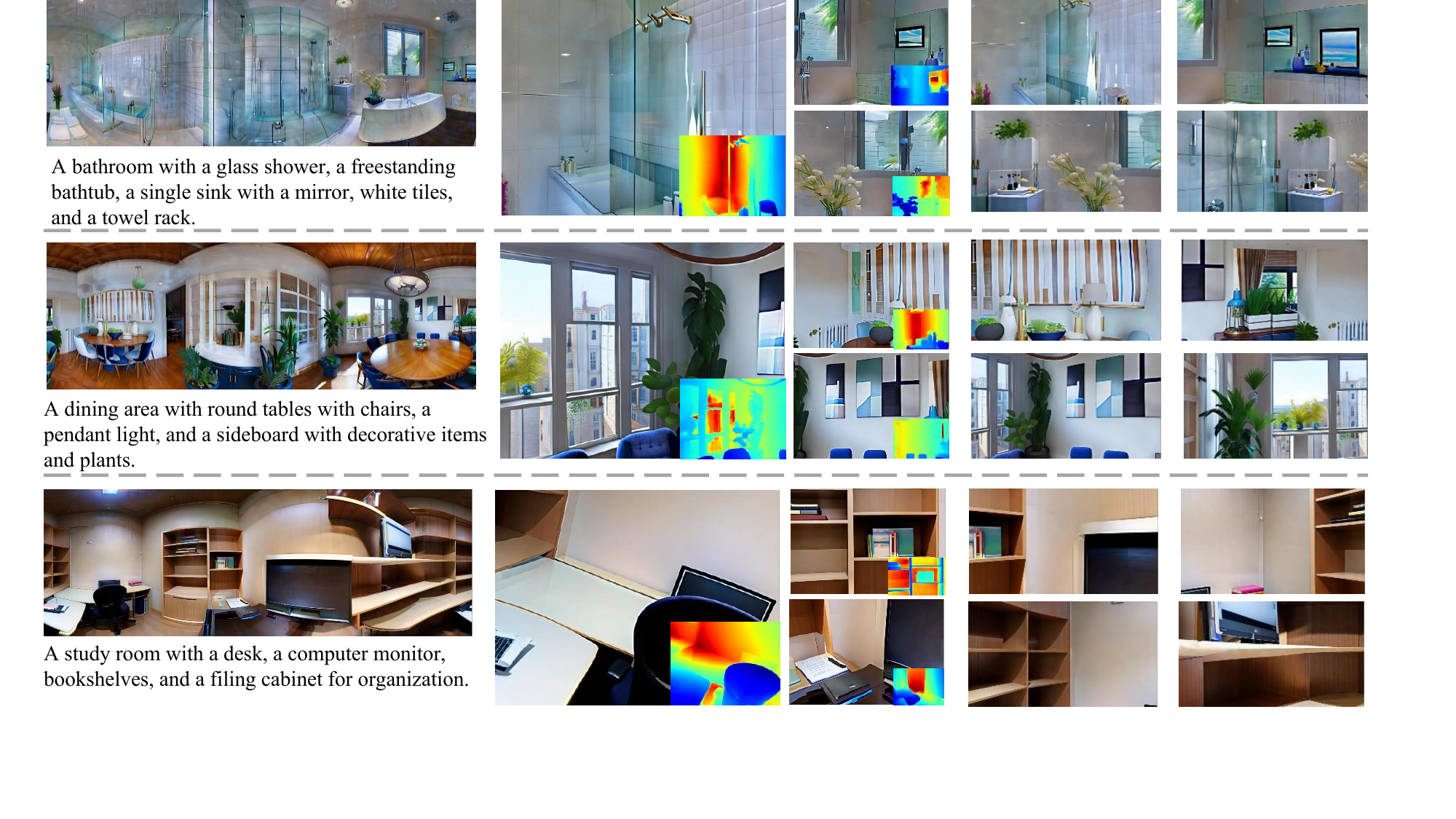}
	\caption{The Panoramic Image on the left represents the high-resolution enhanced panorama generated from a specific text input, showcasing the detailed and cohesive scene captured in 360 degrees. The Multi-view Image in the center displays various perspective images derived from this panorama, illustrating the scene's adaptability across different viewpoints. Finally, the Output on the right presents images rendered from the generated point cloud at new viewpoints, highlighting SceneDreamer360's ability to maintain high quality and spatial consistency in the 3D reconstructions. As shown in the figure, SceneDreamer360 effectively produces detailed, visually coherent, and spatially consistent 3D scenes.}
	\label{visual_fig3}
\end{figure*}
\begin{table*}
\centering
\setlength{\tabcolsep}{12pt} % 调整列间距
\renewcommand{\arraystretch}{1.5} % 调整行间距
% \vspace{0.5em} % 可选：为标题与表格内容之间添加间距
\caption{Comparison of image and render quality across different methods. }
\begin{tabular}{lccccccc}
\hline
\multirow{2}{*}{\textbf{Method}} & \multicolumn{4}{c}{\textbf{Image Quality}} & \multicolumn{3}{c}{\textbf{Render Quality}} \\
\cline{2-8}
 & \textbf{CLIP-Score\textuparrow} & \textbf{Sharp\textuparrow} & \textbf{Colorful\textuparrow} & \textbf{Resolution\textuparrow} & \textbf{PSNR\textuparrow} & \textbf{SSIM\textuparrow} & \textbf{LPIPS\textdownarrow} \\
\hline
Text2Room \cite{hollein2023text2room} & - & - & - & - & 19.29 & 0.728 & 0.206 \\
LucidDreamer \cite{chung2023luciddreamer} & 0.2110 & 0.9437 & 0.3230 & 0.4757 & 29.56 & 0.922 & 0.039 \\
\textbf{SceneDreamer360} & \textbf{0.2534} & \textbf{0.9877} & \textbf{0.3602} & \textbf{0.7082} & \textbf{35.74} & \textbf{0.963} & \textbf{0.035} \\
\hline
\end{tabular}

\label{tab:comparison}
\end{table*}
\begin{table}
\centering
\setlength{\tabcolsep}{9pt} % 调整列间距
\renewcommand{\arraystretch}{1.5}
\caption{Quantitative comparison of different settings. }
\begin{tabular}{lccc}
\hline
Method & PSNR\textuparrow & SSIM\textuparrow & LPIPS\textdownarrow \\ \hline
\textbf{Standard} & \textbf{34.96} & \textbf{0.962} & 0.037 \\
W/o enhance & 30.36 & 0.947 & \textbf{0.011} \\
W/o alignment & 10.30 & 0.365 & 0.732 \\
W/o enhance \& alignment & 9.70 & 0.281 & 0.803 \\
\hline
\end{tabular}

\label{tab:ablation}
\end{table}

\subsection{Baselines} We compare our SceneDreamer360 with the following state-of-the-art baselines:
\begin{itemize}
    \item TextToRoom \cite{hollein2023text2room} employs a mesh-based 3D representation, directly extracting meshes from inpainted RGBD images to represent watertight indoor environments.
    \item LucidDreamer \cite{chung2023luciddreamer} projects outpainted RGBD images onto a point cloud and then supervises the optimization of 3DGS by projecting multiple images from this point cloud. However, since LucidDreamer cannot directly generate 3D scenes from prompts, we utilize a diffusion model to generate condition images, allowing LucidDreamer to create scenes from text input.
\end{itemize}

\textbf{Evaluation Metrics} As there are currently no established evaluation metrics specifically for 3D scene point clouds, we assess our method using metrics adapted from prior work in related areas. We calculate the CLIP-Score \cite{hessel2021clipscore} to measure text-image consistency, allowing us to evaluate how well the generated images align with the input text prompts. To assess the visual quality of single viewpoint renderings, we use CLIP-IQA \cite{wang2023clipIQA}, which provides an image quality evaluation aligned with human perception. Additionally, we employ PSNR, SSIM, and LPIPS to evaluate the fidelity and perceptual quality of the rendered images. These metrics collectively enable a well-rounded assessment, capturing both the accuracy of text-image correspondence and the visual quality of the generated 3D scenes.

\subsection{Qualitative Result}
The point cloud images generated by our method using various prompts are displayed in Fig. \ref{visual_fig3}. These results demonstrate that our method effectively creates 3D spatial scenes that are closely aligned with the input text prompts. Additionally, we show that the generated images can be rendered from arbitrary viewpoint trajectories, highlighting the generalizability and flexibility of our approach. To further evaluate our method, we visually compared the output with LucidDreamer and Text2Room using identical spatial text prompts, as illustrated in Fig. \ref{compare_fig4}. Our method surpasses these alternatives by producing more consistent, complete, and detailed 3D spatial scene point clouds. Specifically, the integration of 3D Gaussian Splatting (3DGS) enables our model to capture finer details and maintain continuous object surfaces, resulting in sharper, more defined images compared to Text2Room, which struggles with clarity and fine details.

Our Stage 1 panoramic optimization also contributes significantly by capturing extensive spatial details, while our tailored rendering trajectory produces comprehensive, cohesive scenes. In contrast, LucidDreamer frequently generates point clouds that are incomplete, with visible voids that compromise the spatial coherence of the scene. Text2Room, on the other hand, often fails to achieve clarity, particularly in regions such as floors and other large surfaces, leading to a lack of detail. In comparison, SceneDreamer360 consistently delivers aesthetically pleasing and well-detailed indoor scenes that accurately reflect the text prompts. The results underscore the superiority of our approach in achieving high-quality, text-consistent, and spatially complete 3D scene reconstructions.

% We quantitatively measured the performance of SceneDreamer360, Text2Room, and LucidDreamer using metrics such as CLIP-Score, CLIP-IQA, PSNR, SSIM, and LPIPS. The results demonstrate that our method outperforms others, highlighting the superiority of SceneDreamer360.

\subsection{Results Comparision}
To objectively assess the quality of the generated images, we employ a range of evaluation metrics, including CLIP-Score, CLIP-IQA, PSNR, SSIM, and LPIPS. Table~\ref{tab:comparison} presents the quantitative results of our evaluation. In terms of generated image quality, SceneDreamer360 consistently outperforms LucidDreamer across multiple metrics, achieving higher scores in CLIP-Score, CLIP-IQA-sharp, CLIP-IQA-colorful, and CLIP-IQA-resolution. These results indicate that our method produces sharper, more colorful, and higher-resolution images that reflect the text prompts with greater fidelity compared to LucidDreamer. Due to the significant perspective variations and incomplete scenes produced by Text2Room, we do not include it in our quantitative image quality measurements. As shown in Fig.~\ref{compare_fig4}, however, the scenes generated by Text2Room display poor consistency and often result in unrealistic or fragmented visual scenarios. This highlights the clear advantage of SceneDreamer360 in generating coherent and realistic multiview 3D scenes.

To evaluate the overall render quality of the 360-degree scenes, we calculate the average values for PSNR, SSIM, and LPIPS across the generated images. SceneDreamer360 consistently outperforms other methods, achieving higher PSNR and SSIM scores, which indicate superior image fidelity and structural similarity. Additionally, our method achieves lower LPIPS scores, reflecting enhanced perceptual quality in the generated images. These findings underscore the effectiveness of our approach in producing high-quality, consistent, and visually appealing 3D multiview images, thereby setting a new benchmark for text-driven 3D scene generation.

\begin{figure}
	\centering
	\includegraphics[width=1\linewidth]{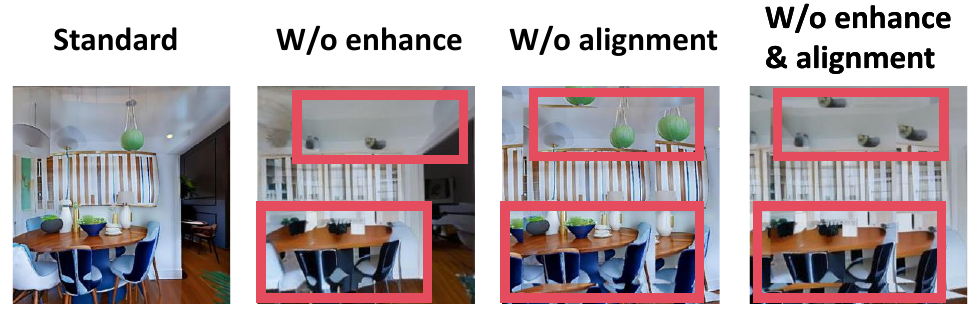}
	\caption{The ablation study of SceneDreamer360.}
	\label{ablation_fig5}
\end{figure}
% Standard represents the full SceneDreamer360 process. W/o enhance indicates the result without the panorama enhancement module. W/o alignment refers to the result without the point cloud alignment module. W/o enhance & alignment shows the result after removing both the panorama enhancement and point cloud alignment modules.

\subsection{Ablation Study}
We conduct ablation experiments on two critical modules—panoramic image enhancement and point cloud alignment—to evaluate their individual contributions to the effectiveness of our method. The primary configurations tested are: “Standard” (using both modules), “W/o Enhance” (without the enhancement module), “W/o Alignment” (without the alignment module), and “W/o Enhance \& Alignment” (without both modules).

As illustrated in Fig.~\ref{ablation_fig5}, the absence of the enhancement module results in a noticeable loss of fine details, indicating that this module is essential for preserving high-resolution features. When the alignment module is removed, we observe significant visual discontinuities, underscoring its role in maintaining spatial coherence across the generated scenes. Notably, the configuration without both modules produces the poorest results, lacking both detail and spatial consistency, which detracts from the realism of the reconstructed 3D scenes.

To further quantify these effects, we measure the performance of the reconstructed images across the different configurations, as shown in Table \ref{tab:ablation}. The results indicate that removing the alignment module leads to a more substantial decline in metrics compared to the removal of the enhancement module, highlighting the alignment module's critical role in ensuring consistent and coherent spatial generation. These findings demonstrate that both modules are integral to our method, with alignment playing a particularly crucial role in achieving high-quality 3D scene reconstruction.

\section{Conclusion}\label{conclusion}
In this study, we introduce a novel text-driven method, SceneDreamer360, which generates 3D panoramic scenes by leveraging large diffusion models, enabling the creation of high-quality scenes without constraints to specific target domains. First, we generate highly coherent panoramic images from input text, producing multi-view consistent, high-quality images that are seamlessly integrated into 3D space as point clouds. Next, we apply 3D Gaussian splatting to further enhance the quality of the 3D scenes. By optimizing panoramic image generation and processing, combined with 3D Gaussian splatting, SceneDreamer360 excels in generating complex 3D scenes with fine-grained details and high resolution. Extensive experiments demonstrate that SceneDreamer360 consistently produces high-quality, diverse 3D scenes under various conditions. Overall, SceneDreamer360 provides an efficient, high-quality solution for text-to-3D scene conversion, contributing to future advancements in 3D scene generation.

% \bibliography{tcsvt}  
% \bibliographystyle{IEEEtran}

% if have a single appendix:
%\appendix[Proof of the Zonklar Equations]
% or
%\appendix  % for no appendix heading
% do not use \section anymore after \appendix, only \section*
% is possibly needed

% use appendices with more than one appendix
% then use \section to start each appendix
% you must declare a \section before using any
% \subsection or using \label (\appendices by itself
% starts a section numbered zero.)
%

% use section* for acknowledgment
% \section*{Acknowledgment}

% Can use something like this to put references on a page
% by themselves when using endfloat and the captionsoff option.

% trigger a \newpage just before the given reference
% number - used to balance the columns on the last page
% adjust value as needed - may need to be readjusted if
% the document is modified later
%\IEEEtriggeratref{8}
% The "triggered" command can be changed if desired:
%\IEEEtriggercmd{\enlargethispage{-5in}}

% references section

% can use a bibliography generated by BibTeX as a .bbl file
% BibTeX documentation can be easily obtained at:
% http://mirror.ctan.org/biblio/bibtex/contrib/doc/
% The IEEEtran BibTeX style support page is at:
% http://www.michaelshell.org/tex/ieeetran/bibtex/

% \bibliographystyle{unsrt}

\begin{IEEEbiography}[{\includegraphics[width=1in,height=1.25in,clip,keepaspectratio]{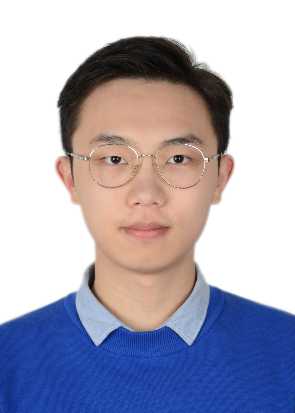}}]{Wenrui Li} received the B.S. degree from the School of Information and Software Engineering, University of Electronic Science and Technology of China (UESTC), Chengdu, China, in 2021. He is currently working toward the Ph.D. degree from the School of Computer Science, Harbin Institute of Technology (HIT), Harbin, China. His research interests include multimedia search, joint source-channel coding, AIGC and spiking neural network.
\end{IEEEbiography}

\begin{IEEEbiography}[{\includegraphics[width=1in,height=1.25in,clip,keepaspectratio]{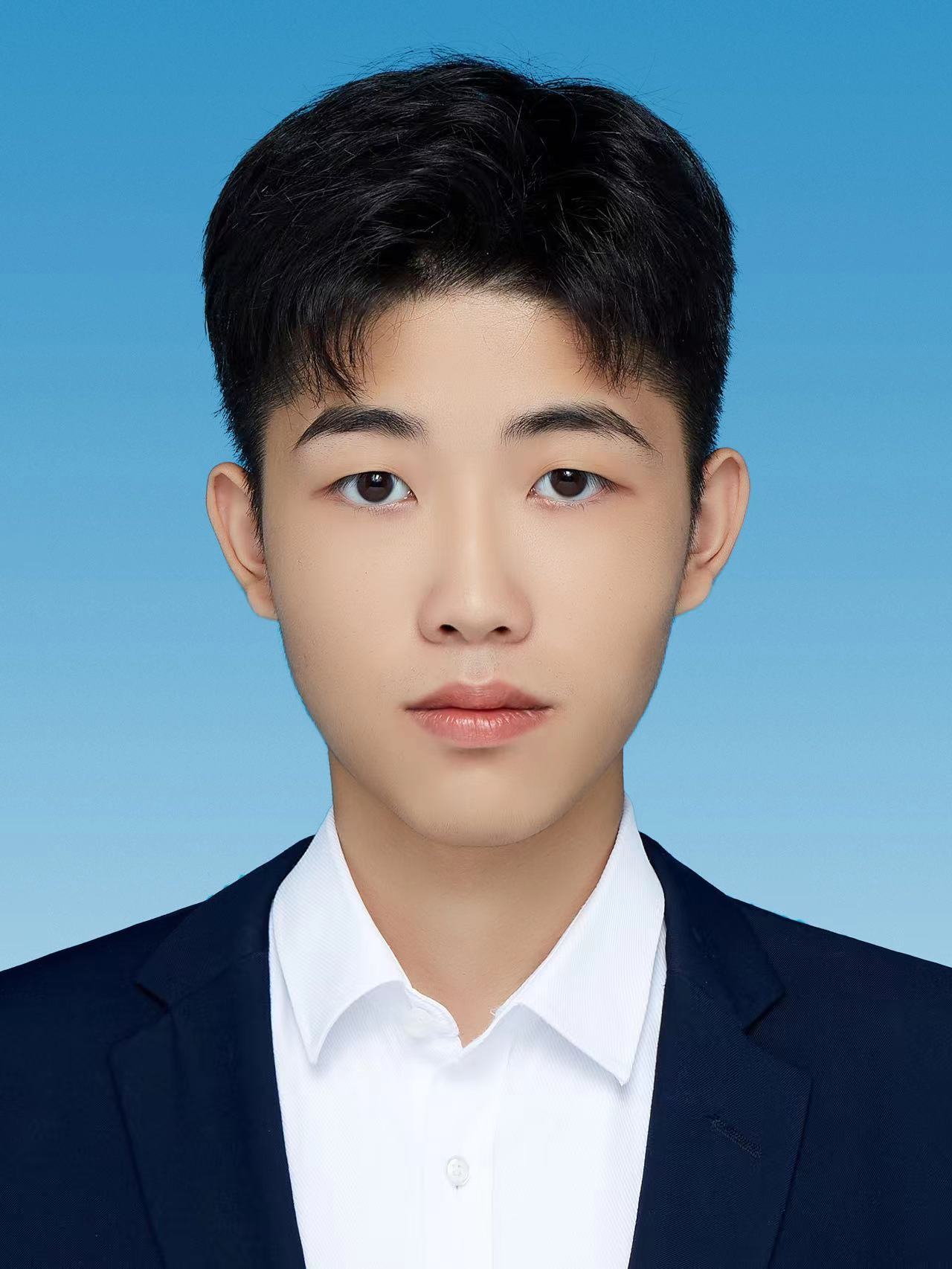}}]{Fucheng Cai} is currently pursuing a Bachelor's degree in Computer Science and Technology at the School of Computer Science and Technology at Harbin Institute of Technology, under the supervision of Professor Fan Xiaopeng. His primary research interests focus on multimodal 3D scene understanding.
\end{IEEEbiography}

\begin{IEEEbiography}[{\includegraphics[width=1in,height=1.25in,clip,keepaspectratio]{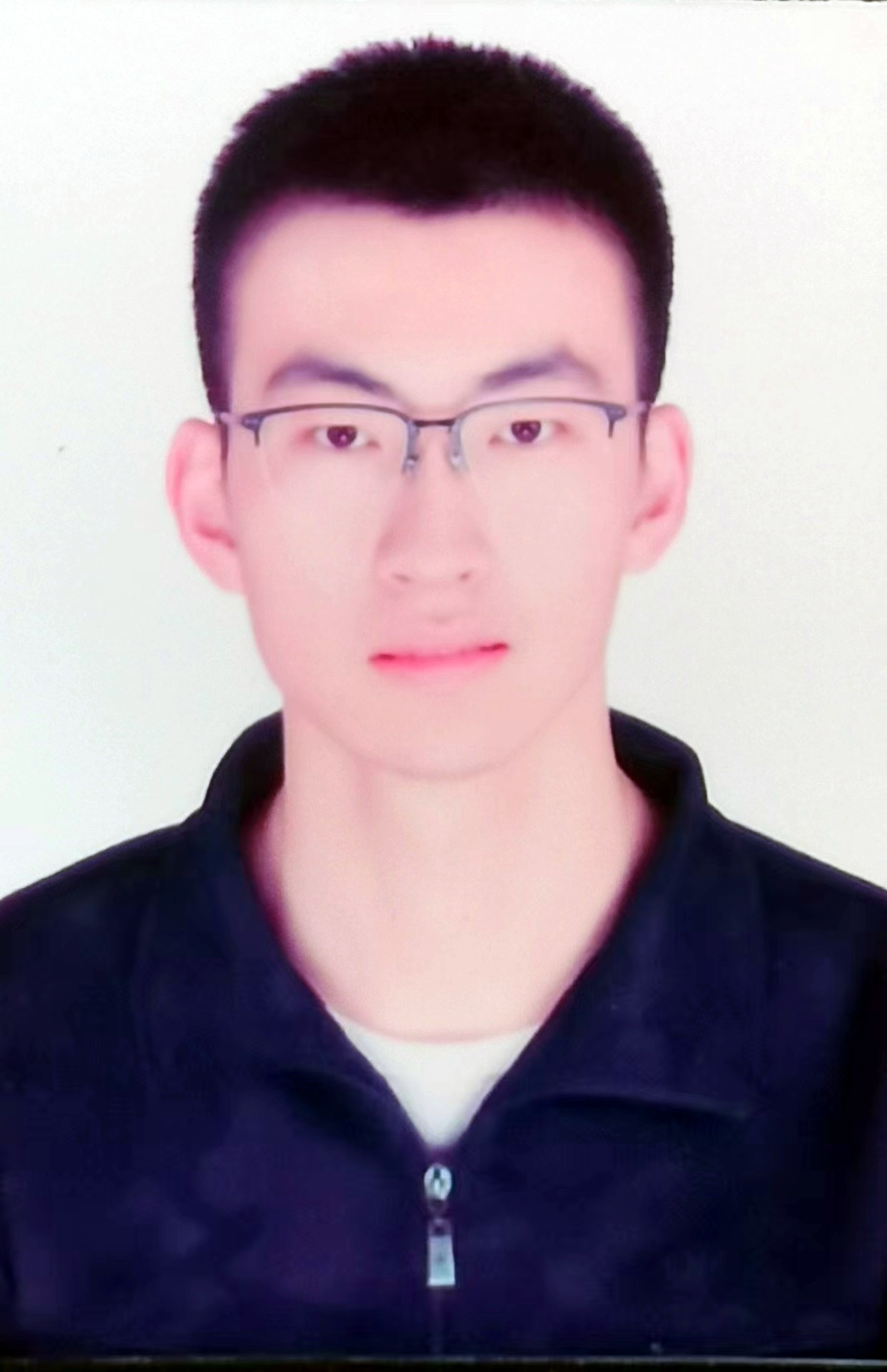}}]{Yapeng Mi} is currently working towards his Bachelor's degree in Computer Science and Technology at Harbin Institute of Technology's School of Computer Science and Technology, under the guidance of Professor Fan Xiaopeng. His main research interests lie in multimodal and 3D scene understanding
\end{IEEEbiography}

\begin{IEEEbiography}[{\includegraphics[width=1in,height=1.25in,clip,keepaspectratio]{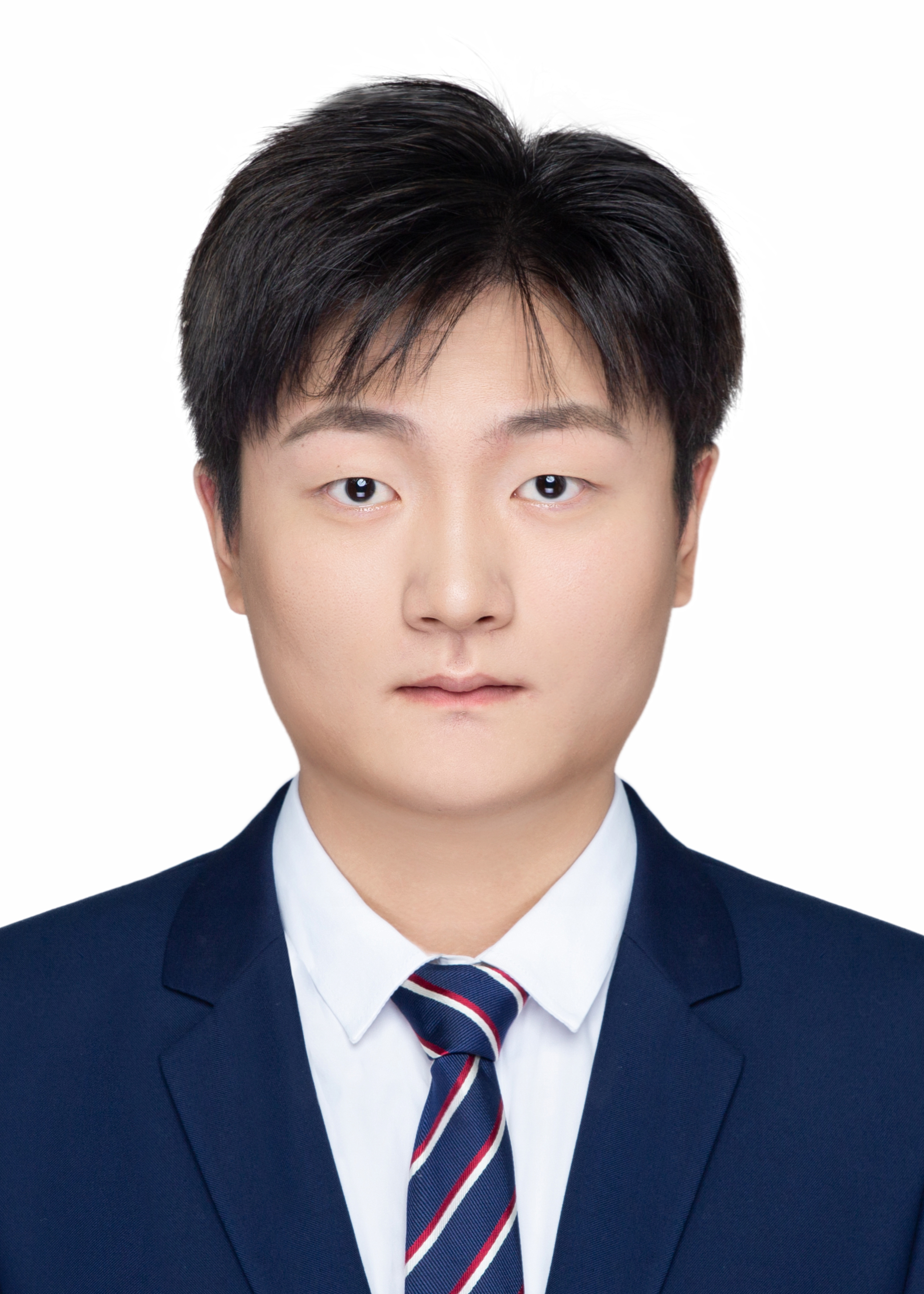}}]{Zhe Yang}  received the B.S. degree from the College of Science, Northeast Forestry University, China, in 2022. He is currently pursuing the M.E. degree with the University of Electronic Science and Technology of China. His research interests include zero-shot learning, audio-visual learning, and computer vision.
\end{IEEEbiography}

\begin{IEEEbiography}[{\includegraphics[width=1in,height=1.25in,clip,keepaspectratio]{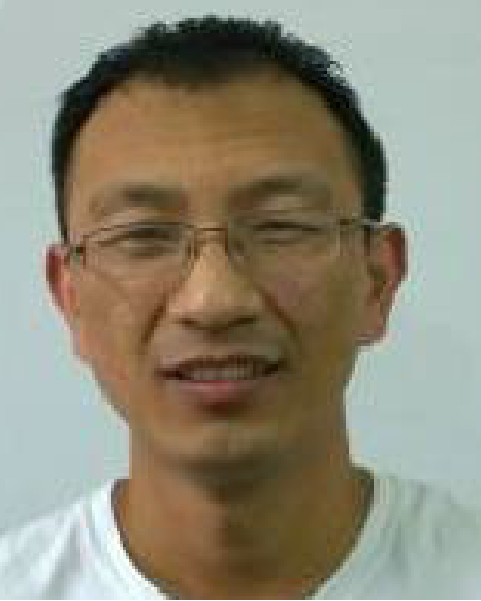}}]{Wangmeng Zuo} (Senior Member, IEEE) received the Ph.D. degree in computer application technology from Harbin Institute of Technology, Harbin, China, in 2007. He is currently a Professor with the Faculty of Computing, Harbin Institute of Technology.
He has published over 200 papers in top tier academic journals and conferences. His current research interests include low level vision, image/video generation, and multimodal understanding. He served as an Associate Editor for IEEE TRANSACTIONS
ON PATTERN ANALYSIS AND MACHINE INTELLIGENCE, IEEE TRANSACTIONS ON IMAGE PROCESSING, and SCIENCE CHINA Information Sciences.
\end{IEEEbiography}

\begin{IEEEbiography}[{\includegraphics[width=1in,height=1.25in,clip,keepaspectratio]{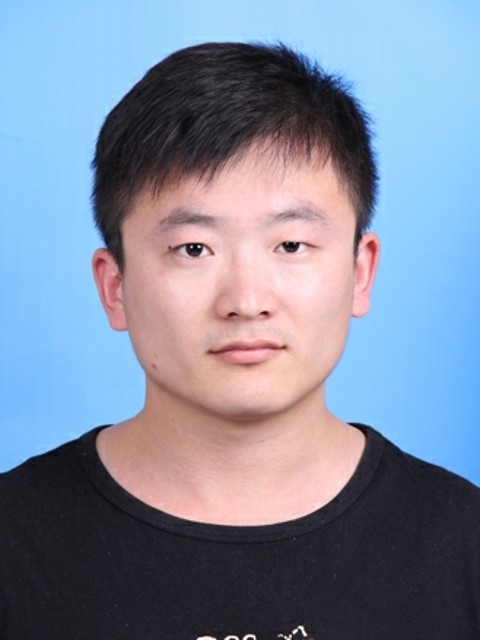}}]{Xingtao Wang} obtained his B.S. degree in Mathematics and Applied Mathematics, as well as his Ph.D. degree in Computer Science, from the Harbin Institute of Technology (HIT) in Harbin, China, in 2016 and 2022, respectively. In 2023, he served as an Assistant Research Fellow at the School of Artificial Intelligence, HIT, and currently holds the position of Associate Researcher. His research focuses on computer graphics, digital twins, and panoramic vision.
\end{IEEEbiography}

\begin{IEEEbiography}[{\includegraphics[width=1in,height=1.25in,clip,keepaspectratio]{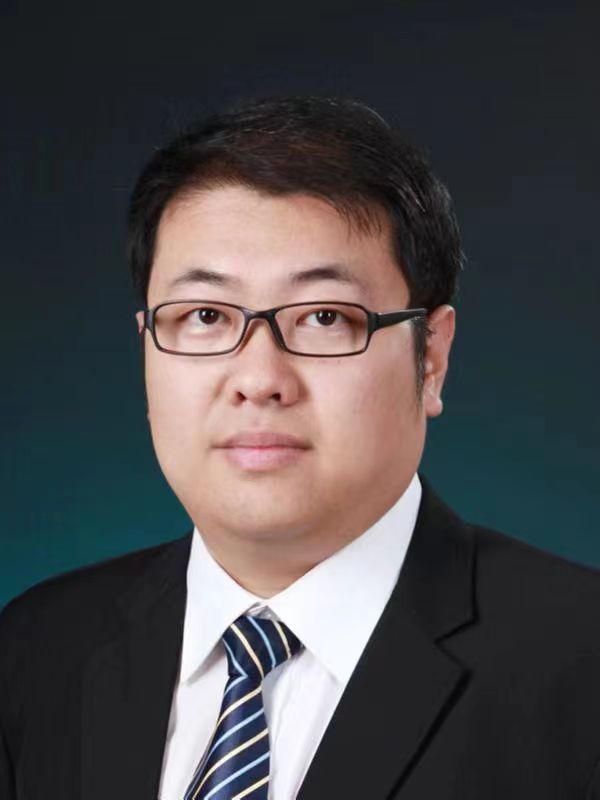}}]{Xiaopeng Fan} (Senior Member, IEEE) received the B.S. and M.S. degrees from the Harbin Institute of Technology (HIT), Harbin, China, in 2001 and 2003, respectively, and the Ph.D. degree from the Hong Kong University of Science and Technology, Hong Kong, in 2009. In 2009, he joined HIT, where he is currently a Professor. From 2003 to 2005, he was with Intel Corporation, China, as a Software Engineer. From 2011 to 2012, he was with Microsoft Research Asia, as a Visiting Researcher. From 2015 to 2016, he was with the Hong Kong University of Science and Technology, as a Research Assistant Professor. He has authored one book and more than 170 articles in refereed journals and conference proceedings. His research interests include video coding and transmission, image processing, and computer vision. He was the Program Chair of PCM2017, Chair of IEEE SGC2015, and Co-Chair of MCSN2015. He was an Associate Editor for IEEE 1857 Standard in 2012. He was the recipient of Outstanding Contributions to the Development of IEEE Standard 1857 by IEEE in 2013.
\end{IEEEbiography}

% % if you will not have a photo at all:
% \begin{IEEEbiographynophoto}{John Doe}
% Biography text here.
% \end{IEEEbiographynophoto}

% % insert where needed to balance the two columns on the last page with
% % biographies
% %\newpage

% \begin{IEEEbiographynophoto}{Jane Doe}
% Biography text here.
% \end{IEEEbiographynophoto}

% You can push biographies down or up by placing
% a \vfill before or after them. The appropriate
% use of \vfill depends on what kind of text is
% on the last page and whether or not the columns
% are being equalized.

%\vfill

% Can be used to pull up biographies so that the bottom of the last one
% is flush with the other column.
%\enlargethispage{-5in}

% that's all folks
\end{document}